\documentclass{article}
\usepackage{ijcai21}

\usepackage{times}
\usepackage{soul}
\usepackage{url}
\usepackage[hidelinks]{hyperref}
\usepackage[utf8]{inputenc}
\usepackage[small]{caption}
\usepackage{graphicx}
\usepackage{amsmath}
\usepackage{amsthm}
\usepackage{booktabs}
\usepackage{bm}
\usepackage{CJKutf8}
\usepackage{graphicx}
\usepackage{multirow}
\usepackage[utf8]{inputenc}
\usepackage{array}
\usepackage{color}

 \usepackage{multirow}

\title{MatchVIE: Exploiting  Match Relevancy between Entities  for Visual Information Extraction}
\author{
	Guozhi Tang$^1$\footnotemark[1],
	Lele Xie$^3$\footnotemark[1],
	Lianwen Jin$^{1,2}$\footnotemark[2],
	Jiapeng Wang$^1$,
	Jingdong Chen$^3$,
	Zhen Xu$^4$,\\
	Qianying Wang$^4$,
	Yaqiang Wu$^4$
	and Hui Li$^4$\
	\affiliations
	$^1$School of Electronic and Information Engineering, South China University of Technology, China\\
	$^2$Guangdong Artificial Intelligence and Digital Economy Laboratory (Pazhou Lab), Guangzhou, China
	$^3$Ant Group, China\\
	$^4$Lenovo Research, China\\
	\emails
	\{eetanggz, eejpwang\}@mail.scut.edu.cn,
	eelwjin@scut.edu.cn, \\
	\{yule.xll, jingdongchen.cjd\}@antgroup.com,
	\{xuzhen8,wangqya,wuyqe,lihuid\} @lenovo.com
}

\begin{document}

\maketitle

\renewcommand{\thefootnote}{\fnsymbol{footnote}} 
\footnotetext[1]{These authors contributed equally to this work.} 
\footnotetext[2]{Corresponding author: Lianwen Jin.} 
\begin{abstract}

Visual Information Extraction (VIE) task aims to extract key information from multifarious document images (e.g., invoices and purchase receipts). Most previous methods treat the VIE task simply as a sequence labeling problem or classification problem, which requires models to carefully identify each kind of semantics by introducing multimodal features, such as font, color, layout. But simply introducing multimodal features  couldn't work well when faced with numeric semantic categories or some ambiguous texts. To address this issue, in this paper we propose a novel key-value matching model based on a graph neural network for VIE (MatchVIE). Through key-value matching based on relevancy evaluation, the proposed MatchVIE can bypass the recognitions to various semantics, and simply focuses on the strong relevancy between entities. Besides, we introduce a simple but effective operation, Num2Vec, to tackle the instability of encoded values, which helps model converge more smoothly. Comprehensive experiments demonstrate that the proposed MatchVIE can significantly outperform previous methods. Notably, to the best of our knowledge, MatchVIE may be the first attempt to tackle the VIE task by modeling the relevancy between keys and values and it is a good complement to the existing methods.
\end{abstract}

\section{Introduction}

 The Visual Information Extraction (VIE) aims to extract key information from document images (invoices, purchase receipts, ID cards, and so on), instead of plain texts. The particularity of the VIE task brings several  additional difficulties. Firstly, documents usually have diverse layouts, which vary significantly even for the same type of the document (e.g., the invoices from different vendors). In addition, the documents may contain multiple similar but not identical texts (e.g., the issue date and the expiration  date) that are very difficult to distinguish.

\begin{figure}[t]
	\centering
		
	\includegraphics[width=1.0\columnwidth]{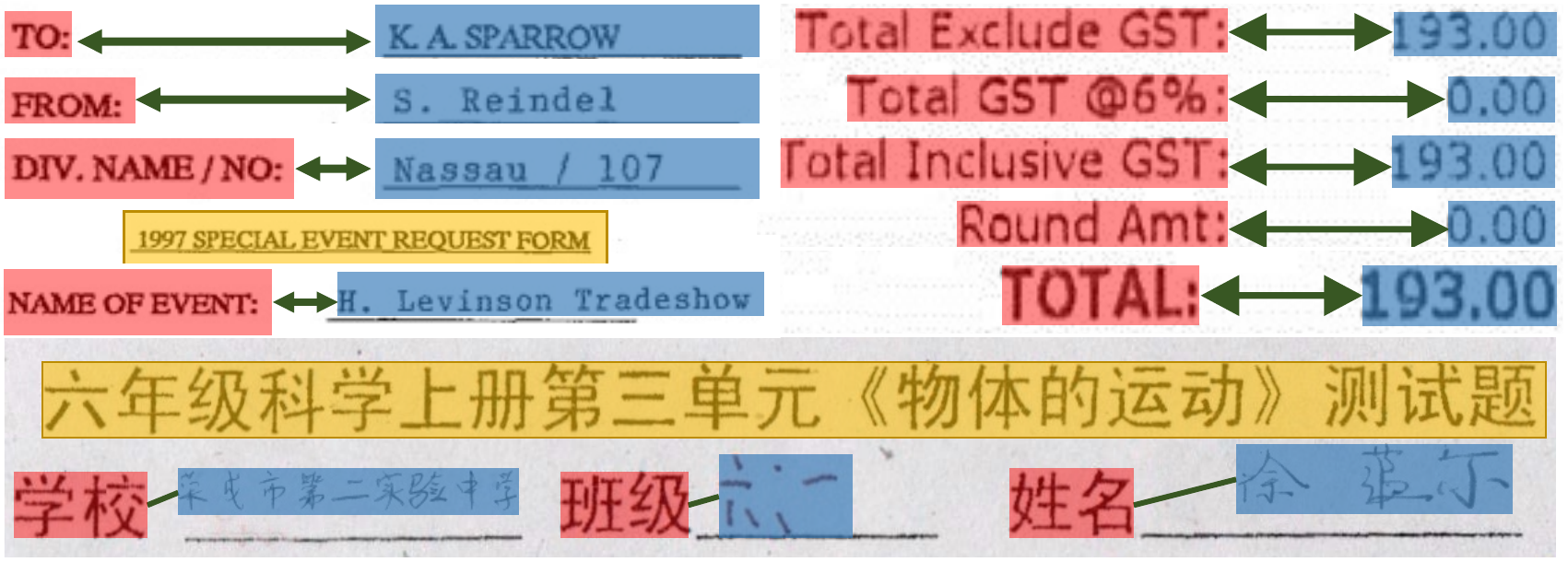}
	\caption{The entity categories of the blue text segments (value) can be identified  according to the semantics of  connected red text segments (key). The entity categories of the yellow text segments can be identified  according themselves semantics. }
	\label{relaveray}
\end{figure}

How to leverage effectively both semantic features and visual features has become the focus of recent studies. Some methods have attempted to incorporate the positions of the texts \cite{jiang2019integrating,hwang2019post}, while others \cite{qian2019graphie,liu2019graph} have modeled the layout information through Graph Neural Network (GNN) \cite{kipf2016semi,velivckovic2017graph}. Some methods \cite{xu2019layoutlm,yu2020pick,zhang2020trie} utilized the Convolutional Neural Network (CNN) \cite{he2016deep,huang2017densely} to fetch image features and fused them with semantic features for improving performance. These methods have achieved improved results by considering the visual features. However, most previous methods are confined to sequence labeling or direct classification, which requires models to assign each entity with corresponding labels carefully when facing numerous semantic categories or some ambiguous texts. In visual-rich documents, the layout information between entities is an important reasoning clue.   As shown in Figure 1, the entity categories of \textbf{ blue text segments (value)} can be identified  according to the semantic of \textbf{red text segments  (key)}.

Compared with sequence labeling or direct classification, we find that studying the relevancy between keys and values could be an another effective solution for the VIE, based on the following observations and considerations: \textbf{(1)} The texts in document images usually appear in the form of key-value pairs.  If the corresponding key can be found for a specific value, the attribute (category) of this value can be naturally determined. \textbf{(2)} There may be multiple similar texts in one document image (e.g., registration, amount and expiry dates), and keys of these values can help the model distinguish among them. \textbf{(3)} Considering the relevancy between keys and values could significantly simplify the learning process for models and bypass the recognition to similar  semantic. \textbf{(4)} As for the standalone texts (without keys), they are easy to identify by semantics. This is also why these values can appear independently in the document images.

Therefore, in this paper we propose a novel key-value matching model for VIE (MatchVIE). It  can effectively integrate the semantic, location and visual information of entities and innovatively consider the edge relationship of the graph network to evaluate the relevance of entities. There are two branches in the network, one is to measure the relevancy between keys and values, and the other auxiliary branch is to process the standalone texts by combining  sequence labeling.  To the best of our knowledge, the proposed MatchVIE may be the first attempt to accomplish the VIE task by modeling the relevancy between keys and values. The main contributions of our study are summarized as follows:
\begin{itemize}
\setlength{\itemsep}{0pt}
\setlength{\parsep}{0pt}
\setlength{\parskip}{2pt}
\item Our MatchVIE  effectively incorporates features (including layout, entity relevance, etc.) and brings significant accuracy improvements and surpasses  existing methods.
\item Our MatchVIE  bypasses the recognition to various similar semantics by  focusing only  on the strong relevancy of entities. This simplifies the process of VIE.
\item We introduce a simple yet effective operation named Num2Vec to tackle the instability of encoded values, which helps model converge more smoothly. 
\item Most importantly, this paper demonstrates that the visual information can be effectively extracted through modeling the relevancy between keys and values, which provides a brand-new perspective to solve the VIE task.
\end{itemize}

\section{Related Works}

Recently, VIE has attracted the attention of many researchers.  Researchers are of the view that  image features of  documents are very useful because features are mixed representations of fonts, glyph, colors, etc.   As the VIE task involves the document images, some researchers regard it as a pure computer vision task, such as EATEN\cite{guo2019eaten},  TreyNet\cite{carbonell2020neural} and \cite{borocs2020comparison}.  These methods solved the VIE task from the perspective of Optical Character Recognition (OCR ).  For each type of entities, these methods designed corresponding decoders that were responsible for recognizing the textual content and determining its category.   This method couldn't work well when faced with complex layouts  because of the absence of  semantic features.

 The core of the research is how to make full use of the multimodal  features of  document images.  Researchers have approached VIE from various perspectives. \cite{hwang2019post} and  \cite{jiang2019integrating} serialized the text segments based on the coordinate information and fed coordinates to a sequence tagger. However, simply treating the position as some kind of features might not fully exploit the visual relationships among texts.  To make full use of the semantic features and position information, Chargrid \cite{katti2018chargrid} mapped  characters to one-hot vectors which filled the character regions on document images. An  image with semantic information is  fed  into a CNN for detection and semantic segmentation to extract entities. The later BERTgrid \cite{denk2019bertgrid} followed a similar approach but utilized different word embedding methods. However, it  introduced a vast amount of calculation by  using channel features to represent semantics, especially languages with large categories.

  Therefore, it is usually a better solution to construct a global document graph, using semantic features as node features and the spatial location features of text segments as edge features. Several methods \cite{qian2019graphie,liu2019graph,yu2020pick,gal2020cardinal,cheng2020one} employed the GNN to model the layout information of documents. Through the messages passing between nodes, these models could learn the overall layout and distribution of each text, which was conductive to the subsequent entity extraction. For example, \cite{gui2019lexicon} proposed a lexicon-based graph neural network that treated the Chinese NER (Named Entity Recognition) as a node classification task. Besides, the GraphIE \cite{qian2019graphie} and the model proposed by \cite{liu2019graph} extracted the visual features through GNN to enhance the input of the BiLSTM-CRF model, which was proved to be effective. Different from the fully-connected or hand-crafted graph, the PICK \cite{yu2020pick} predicted the connections between nodes through graph learning \cite{jiang2019semi}, which also boosts the results.  These methods used the GNN to encode text embeddings given visually rich context  to learn the key-value relationship implicitly. It is difficult to ensure that models can learn it well. However, our method explicitly learns key-value matching which makes full use of  edge features for relevancy evaluation. 

\section{Methodology}

The framework of  the proposed MatchVIE is presented in Figure 2.  It consists of a multi-feature extraction backbone and two specific branches of relevancy evaluation and entity recognition, respectively.  The multi-feature extraction backbone considers features (e.g.  position, image, and semantics)  all together.  Then the  relevancy evaluation branch is based on a graph module to model the overall layout information of documents and obtain  the key-value matching probability. Meanwhile,  to solve the  sequence labels of standalone text segments, we  design an entity recognition branch. 

\begin{figure*}[h]
	\centering
	\label{overall framework}

	\includegraphics[width=1.8\columnwidth]{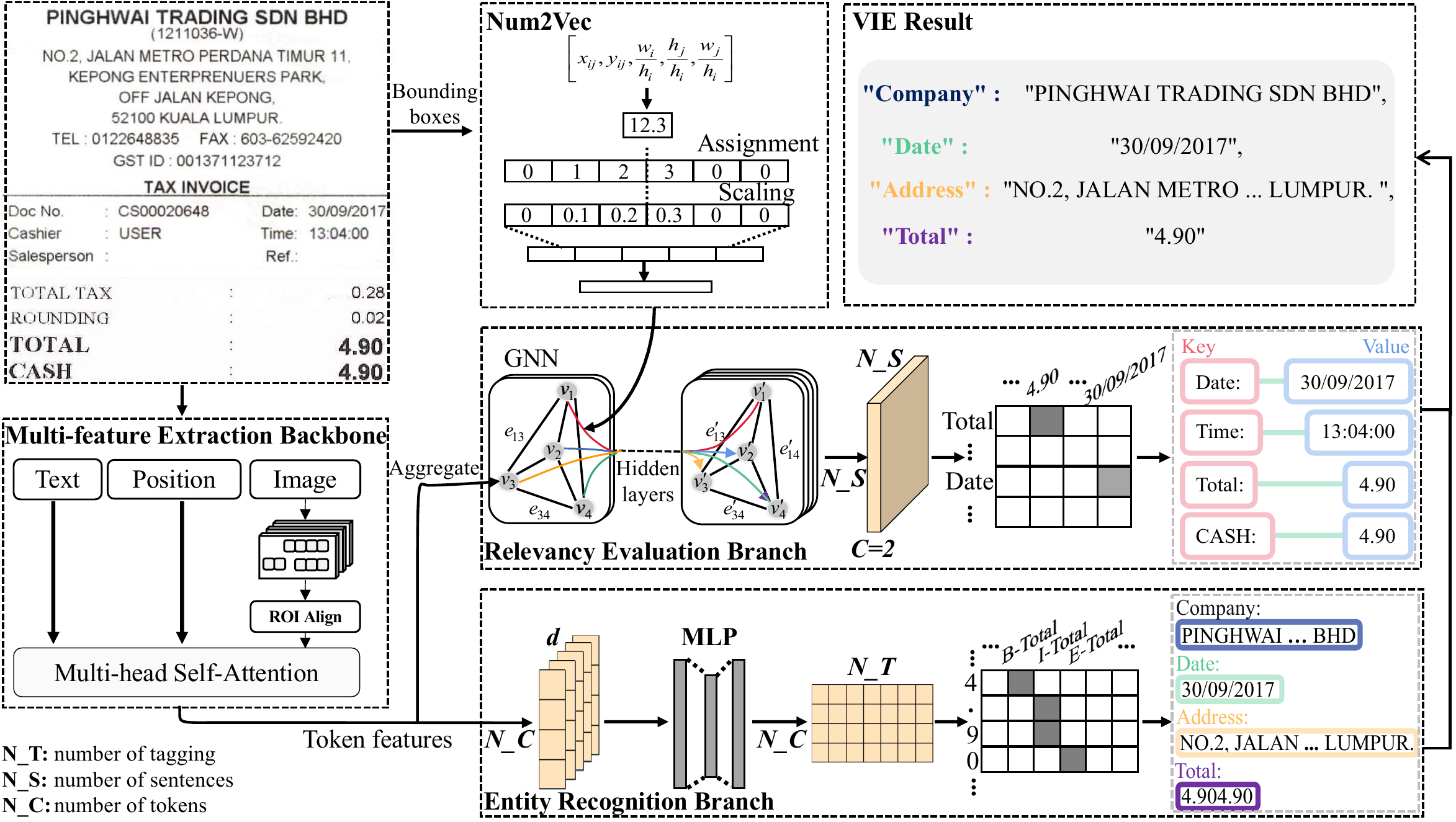}

	\caption{Overall framework of MatchVIE. The relevancy evaluation branch predicts the key-value relationship between entities. The entity recognition branch mainly determines the categories of standalone  entities. The entity recognition branch is difficult to distinguish numeric categories which are similar in visual and semantic, such as the  `4.90' in purple.}
\end{figure*}

\subsection{Multi-feature Extraction Backbone}
 To explore the features of multimodal, we propose  an effective way to consider  position features, visual features and textual features.  We consider the input to be a set of text segments. As for textual features, given the $i$-th text segment $T_{i} = (t_{1}^i,t_{2}^i,t_{3}^i,...,t_{L}^i)$, where $L$ is the number of tokens in it, and $t_{m}$ is the $m$-th token embedding in the text segment.  Inspired by \cite{devlin2019bert}, we apply the pre-train model to initialize the parameters.  As for  position embedding, different from the position embedding that represents the word sequence, we use the spatial position of each token in the document.   In detail, given a  position embedding $P_{i} = (p_{1},p_{2},p_{3},...,p_{L})$, the position embedding $p_{m} $ for $m$-th token is obtained by linearly  transforming its bounding box coordinates to a $d$-dim vector.

 As for image visual embedding, previous studies \cite{yu2020pick} usually applied a  CNN for catching  morphology information from  each cropped image. This is unable to capture global visual information because of isolating each text segment. Therefore, we  attempt to extract the features from the whole image by ResNet \cite{boroumand2018deep}. Then, the ROI-Align \cite{he2017mask} is used to obtain the  Region of Interest (ROI) according to the coordinate of tokens.  Similarly,   the visual embedding of the $i$-th text segment is denoted by  $I_{i} = (i_{1},i_{2},i_{3},...,i_{L})$. The $m$-th  token visual embedding is given as follows:
  \begin{equation}
I_{m} = ROIAlign(ConvNet(X)),
\end{equation}
where $X$ is the inputted  image. In order to capture local and non-local features, we use the self-attention mechanism to obtain the context features  $\hat{C}$.  Afterwards, we obtain the $Q$,$K$, $V$ in the  scaled dot-product attention by combining the  position features, visual features and textual features together. In detail, the context feature  $\hat C$ is obtained by,
\begin{equation}
	\begin{split}
	\hat C&= MultiHead(Q,K,V) \\
	&=[head_{1}, head_{1},...,head_{n}]W^{d},\\
	\end{split}
\end{equation}
\begin{equation}
Q,K,V = LayerNorm(Linear(I)+P+T),
\end{equation}
where $T$, $P$ and $I$  is the textual embedding, position embedding and  visual embedding, respectively. The $d$ is the embedding dimension  intended to be  768.

\subsection{Relevancy Evaluation Branch}
To represent the relevancy  between entities, the layout information such as the distance between text segments is an important clue as demonstrated in \cite{cheng2020one}.  We construct a fully-connected graph for the document,  which  means each node is connected to each other. Given a graph $G = (V, E)$, $v_i \in V (i=1,2,...,N)$ denotes one of the $N$ nodes (text segments) in a graph document, and $e_{ij} \in E$ represents the edge between node $v_i$ and $v_j$.

The initial node features are obtained by aggregating the context feature  $\hat C$  in units of  text segments. The initial edge features are expected to represent the relationships in vision between text segments. Therefore, we encode the visual information into the edges.  Previously, \cite{liu2019graph} defined edge features as,
\begin{equation}
	\label{edge features}
	\bm{{\rm{e}}}_{ij}=[x_{ij}, y_{ij}, \frac{w_i}{h_i}, \frac{h_j}{h_i}, \frac{w_j}{h_i}],
\end{equation}where $x_{ij}$, $y_{ij}$ represents the relative position between node $v_i$ and $v_j$; $w_i/h_i$ denotes the aspect ratio of node $v_i$.  However, we find that the encoded numeric values are very unstable  because of the diversity of  distance and shape of different  text segments.  To address this issue, we propose a simple yet effective operation, namely Num2Vec to process each item. As shown in Figure 2, for a numeric value, we provide a fixed-length of 8 arrays to hold each digit position. The first half of the array corresponds to an integral part, and the rest corresponds to the fractional part. Then, we scale these digits by a factor of 0.1.  Consequently, the encoded values are constrained to the range of $[-0.9,+0.9]$, which can effectively reduce the fluctuation range of the data.

In the process of features update in the GNN, these two types of features are mixed together for later predictions. We follow the approach adopted by \cite{liu2019graph} who defined a triplet $(\bm{{\rm{v}}}_i,\bm{{\rm{e}}}_{ij},\bm{{\rm{v}}}_j)$ for features updating. The triplet feature is linearly transformed by a learnable weight $\bm{{\rm{W}}}_g$, which generates an intermediate feature $\bm{{\rm{g}}}_{ij}$,
\begin{equation}
\label{intermediate feature}
\bm{{\rm{g}}}_{ij}=\bm{{\rm{W}}}_g[\bm{{\rm{v}}}_i||\bm{{\rm{e}}}_{ij}||\bm{{\rm{v}}}_j].
\end{equation}

Then, we  apply $\bm{{\rm{g}}}_{ij}$ to update the edge features through another linear transformation (applying a nonlinearity, $\sigma$):

 \begin{equation}
\label{update edge}
\bm{{\rm{e}}}_{ij}{'}=\sigma(\bm{{\rm{W}}}_e\bm{{\rm{g}}}_{ij}),
\end{equation}where the $\bm{{\rm{g}}}_{ij}$ is also used to evaluate the attention coefficients and the new node features. The $\bm{{\rm{e}}}_{ij}{'} $ is the intermediate update state of edge features. We also used multi-head attention \cite{velivckovic2017graph} to enhance the feature learning.
\paragraph{Relevancy Evaluation.}
The combination of text segments can be enumerated by $N^2$ edges. We model the relevancy evaluation as a binary classification problem. If two text segments form a key-value pair, the combination is treated as positive; otherwise, it is negative. In detail, we feed the last edge features $\bm{{\rm{e}}}_{ij}$  to another multi-layer perceptron (MLP) which predicts two logits for each edge.  After performing softmax activation on the logits, we acquire a matching probability regarded as relevancy between two text segments.  When training this branch, it is notable that the matching matrix of size $N \times N$ is very sparse for the correct combinations. The loss of positive samples is easily overwhelmed by the negative ones. To solve  this problem, the loss function for the relevancy evaluation branch is designed as the focal loss \cite{lin2017focal} which can automatically adjust weight coefficients based on the difficulty of samples. The procedure is as follows:

\begin{equation}
\label{match branch}
\bm{{\rm{p}}}_{ij}={Softmax}({\rm MLP}(\bm{{\rm{e}}}_{ij})),
\end{equation}

\begin{small} 
	\begin{equation}
			\mathcal{L}_{Re}(p,y^*) = \left\{
		\begin{split}
			-\alpha(1-p{'}_{ij})^\gamma log(p{'}_{ij})&,\quad y_{ij}^*=1, \\
			-(1-\alpha)p{'}_{ij}^{\gamma} log(1-p{'}_{ij})&,\quad otherwise,
		\end{split}
		\right.
	\end{equation}
\end{small}where $\bm{{\rm{p}}}_{ij}$ is the predicted probability vector and the $p_{ij}^{'}$ represents the probability for the positive class. The $y_{ij}^*$ is the label for edge $e_{ij}$. The $\gamma$ is a focusing parameter intended to be  2. The $\alpha$ is used to balance the positive and negative classes and we set it to 0.75 in our experiments.

\subsection{Entity Recognition Branch}

Some standalone text segments are usually easy to distinguish by semantics, but their  attributes  couldn't be  determined without key-value pairs. To address this issue, we design  the entity recognition branch. Note that this branch is oriented to standalone text segments. The implementation for this branch is not unique, here we follow common sequence labeling methods.  We feed the context feature $\hat{C}$ to a MLP, projecting the output to the dimension of BIOES tagging \cite{sang1999representing} space.  For each token, we perform element-wise sigmoid activation on the output vector to yield the predicted probability of each class,

\begin{equation}
		P_{entity}  = Softmax({\rm MLP}(\hat{C})).
\end{equation}
Then a CRF layer is applied to  learn the semantics constraints in an entity sequence. For CRF training, we minimize the negative log-likelihood estimation of the correct entity sequence and calculate the loss 	$\mathcal{L}_{entity}$.

\subsection{Two-branch Merge Strategy}
 During training, the proposed two branches can be trained and the losses are generated from two parts,
  \begin{equation}
 	\mathcal{L}  = \lambda_{entity}\mathcal{L}_{entity} +   \lambda_{Re}\mathcal{L}_{Re}, 
 \end{equation}
where hyper-parameters  $\lambda_{entity}$ and  $\lambda_{Re}$ control the trade-off between losses, we set them to 1.0.
 During inference, It is worth to note that we give priority to the classification results of the relevancy evaluation branch.  In relevancy evaluation branch,  if the matching probability is higher than  a threshold (0.5), these two text segments are regarded as a key-value pair. For  remaining texts which are considered as standalone texts, prediction results of the entity recognition branch are adopted.  The attribute (category) of a value can be determined according to its corresponding key. A simple approach is to perform lookup on a mapping table that contains all possible keys for a certain category. The mapping table needs to be created in advance based on training set. It is not flexible to customize all the categories. Therefore, we propose another way to determine the category based on the semantic similarity. The details are as follows:

\begin{footnotesize} 
	\begin{equation}
		\begin{aligned}
			\label{semantic similarity eq}
			\rm{\textbf{Z}}\!=\mathop{min}\limits_{j\in\{C\}}\Vert{\rm BiLSTM}(\bm{{\rm{x}}}_{1:m}^i;\bm{{\rm{\Theta}}}{'})\!-{\rm BiLSTM}(\bm{{\rm{y}}}_{1:n}^j;\bm{{\rm{\Theta}}}{'})\Vert_2,
		\end{aligned}
	\end{equation}
\end{footnotesize}where $\bm{{\rm{x}}}_{1:m}^i$ are word embedding for each token in the $key$. Similarly, the $\bm{{\rm{y}}}_{1:n}^j$ are those for the $category$ name. We use the pre-trained models to initialize the word embedding features (semantic features) of texts.  To acquire the semantic representation, we input the $key$ and $category$ name to an off-the-shelf BiLSTM model to get the context encoding over all tokens. The $\bm{{\rm{\Theta}}}{'}$ is the learnable weight of BiLSTM which is fixed during inference. We then compute the L2 distance for the semantic feature vectors of the $key$ and $category$ name. The one with the minimum distance is selected as the category of the key-value pair.

 \section{Experiments}
   \subsection{Datasets}
 We conduct experiments on three real-world public datasets, whose statistics are given in Table 1.   Note that the layout of the three datasets is variable. 

\begin{table}[!htb]
	\centering
	\setlength{\tabcolsep}{2mm}{

		\begin{tabular}{lccccc}
			\hline
			\textbf{Dataset} & \textbf{Training} & \textbf{Testing} & \textbf{Entities}  & \textbf{K-V Ratio(\%)} \\ \hline
			FUNSD           & 149               & 50               & 4                           & 74.69              \\
			EPHOIE          & 1183              & 311              & 10                        & 85.62              \\
			SROIE           & 626               & 347              & 4                         & 54.33              \\ \hline
			
	\end{tabular}}
	\caption{Statistics of datasets used in this paper, including the division of training/testing sets,  the categories of named entities, and the proportion of key-value pairs.}
	\label{tab1}
\end{table}

 FUNSD\cite{jaume2019funsd} is a public dataset of 199 fully annotated forms, which has  4 entities to extract (i.e., Question, Answer, Header and Other).
 
  EPHOIE\cite{wangjia}  is a public  dataset that consists of 1,494 images of Chinese examination paper head which has 10  entities to extract (i.e., Subject, Name, Class).
  
   SROIE\cite{huang2019icdar2019} is a public dataset that contains 973 receipts in total. Each receipt is labeled with 4 types of entities (i.e., Company, Address, Date, and Total).

      \subsection{Implementation Details}
 The model is trained from using the Adam optimizer with a learning rate of 0.0005 to minimize the 	$\mathcal{L}_{entity}$ and 	$\mathcal{L}_{Re}$  jointly during the training.  The feature extractor for catching image features is implemented by  ResNet-34 \cite{boroumand2018deep}. We apply the BIOES tagging scheme for entity recognition branch. In relevancy evaluation branch, we set the number of graph convolution  layers to 2 and 8 heads for the multi-head attention. We  use the text-lines which have already been annotated in the datasets as the text segments.

\subsection{Ablation Study}

\begin{table}[t]
	\centering
	\setlength{\tabcolsep}{2mm}{
		
		\begin{tabular}{lccccc}
			\hline
			\multirow{2}{*}{\textbf{Module}} & \textbf{FUNSD} & \textbf{} & \textbf{EPHOIE} & \textbf{} & \textbf{SROIE} \\ \cline{2-2} \cline{4-4} \cline{6-6} 
			& \textbf{F1(\%)}    & \textbf{} & \textbf{F1(\%)}     & \textbf{} & \textbf{F1(\%)}    \\ \hline
				\textbf{MatchVIE(Ours)}                         & \textbf{81.33}          &           & \textbf{96.87}        &           & \textbf{96.57}          \\
			(-)Focal Loss                    & 80.66         &           &  96.28           &           & 95.21         \\
			(-)K-V matching                  & 76.47          &           &  92.19          &           & 93.23          \\
			(-)Num2Vec                       & 74.25          &           & 90.31          &           & 91.31          \\ \hline
	\end{tabular}}
	\caption{Ablation study (F1-score) of the proposed model on the three datasets.   }
	\label{ab1}
\end{table}
\begin{figure}[t]
	\centering

	\includegraphics[width=1.0\columnwidth]{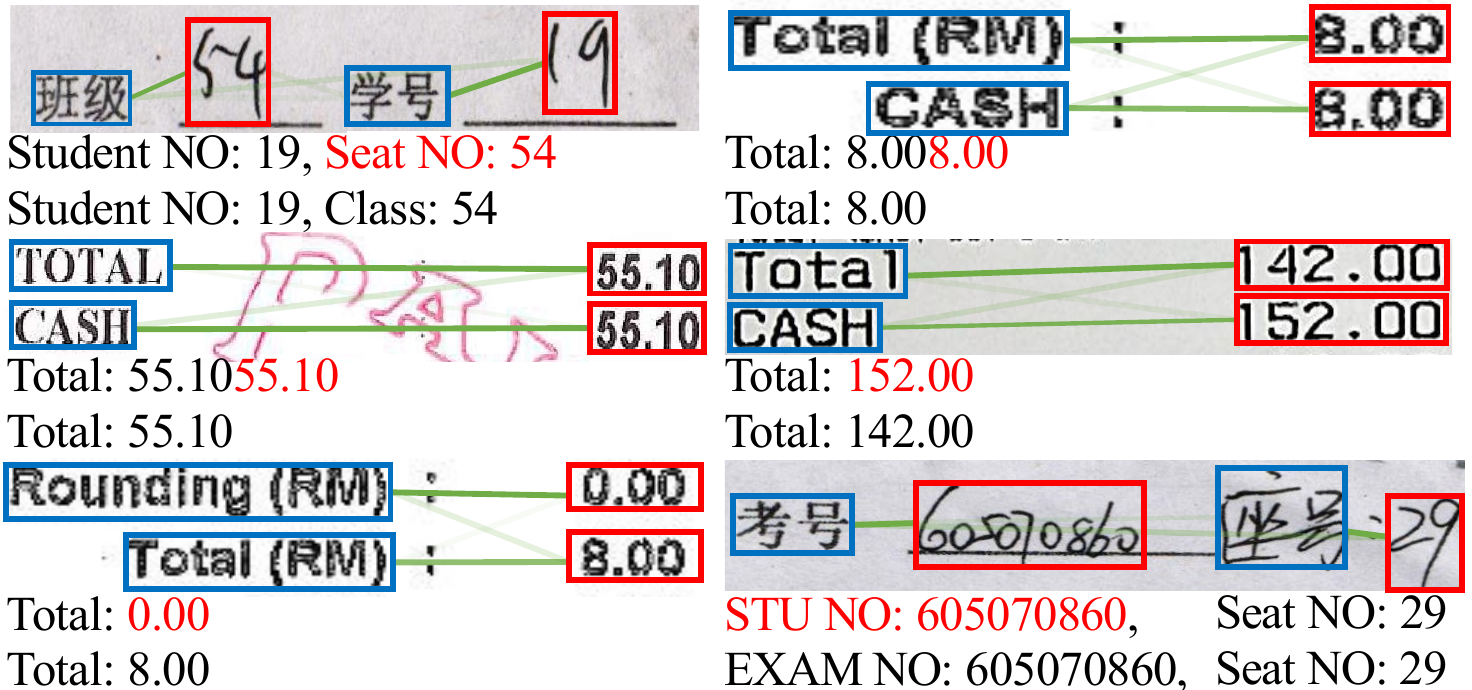}

	\caption{Comparison between the prediction results  of MatchVIE (\textbf{bottom row}) and sequence labeling methods (\textbf{top row}). Red blocks: $value$, blue blocks: $key$, red font: error prediction, green line depth: matching confidence level.}
	\label{relaveray}
\end{figure}

	\begin{table}[t]
	\centering
	\label{ab5}
	
	\setlength{\tabcolsep}{1.0mm}{
		\begin{tabular}{ccccccc}
			\hline
			\multirow{2}{*}{\textbf{Setting}} & 	\multirow{2}{*}{\textbf{Meathods}} & \textbf{FUNSD}        & \textbf{} & \textbf{EPHOIE}       & \textbf{} & \textbf{SROIE}        \\ \cline{3-3} \cline{5-5} \cline{7-7} 
			&                                                                                                                         & \textbf{F1(\%)} & \textbf{} & \textbf{F1(\%)} & \textbf{} & \textbf{F1(\%)} \\ \hline
			1)                                & Lookup table                                                                                                            & 80.44             &           & \textbf{96.96}               &           & 96.27                \\
			Ours                              & Semantic similarity                                                                                                     & \textbf{81.33}                 &           & 96.87                &           & \textbf{96.57}                 \\ \hline
	\end{tabular}}
	\caption{Two methods of mapping keys to certain categories.}
\end{table}

\begin{figure}[t]
	\centering
	\includegraphics[width=0.95\columnwidth]{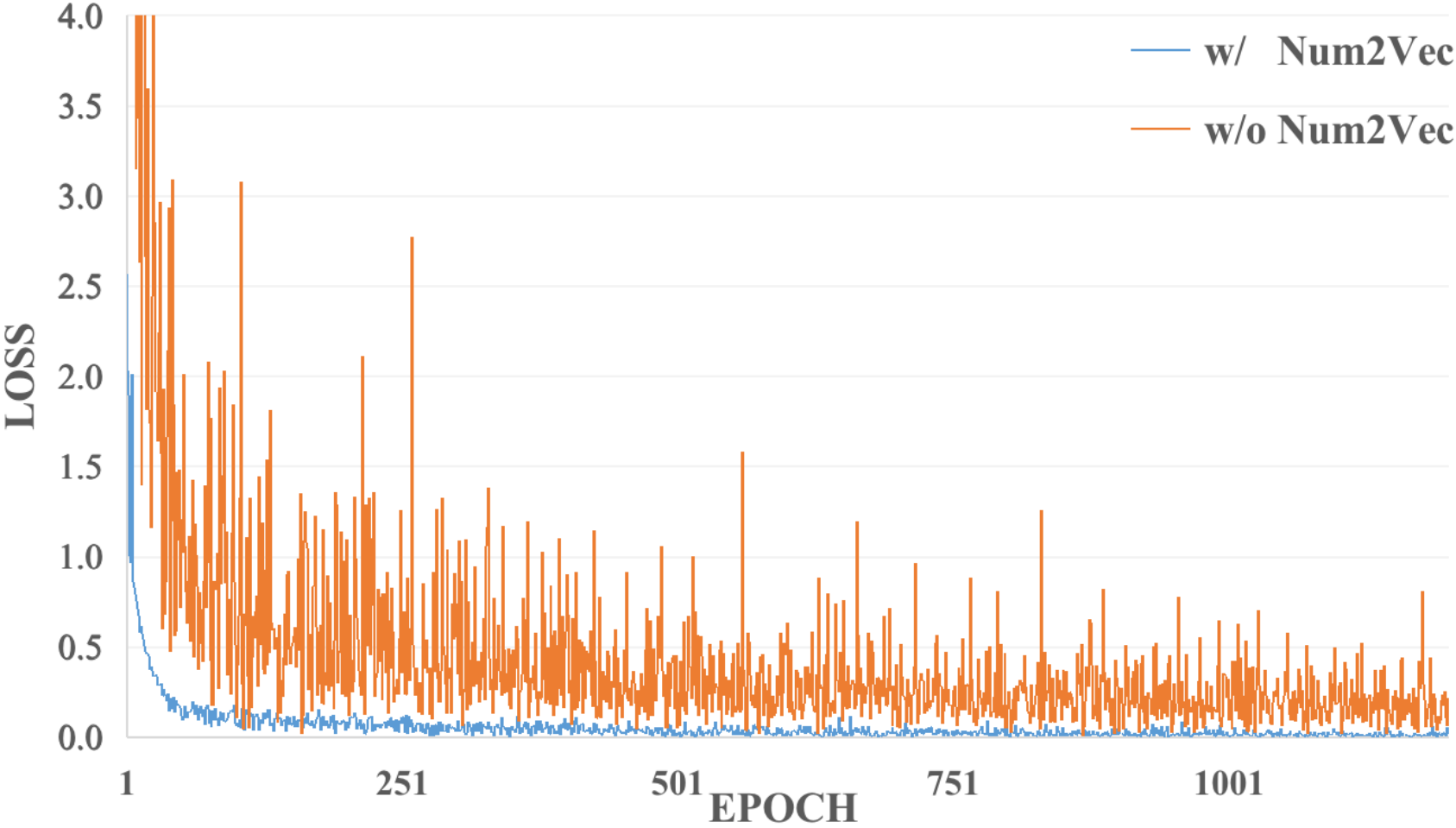}
\caption{The yellow polyline indicates loss  without Num2Vec and the blue polyline indicates the loss with Num2Vec. It can be seen from that using Num2Vec can make training converge smoothly. }
\label{lossnum2vec}

\end{figure}

 As shown in Table 2, we analyze the effect of each component in MatchVIE, including the focal loss, K-V matching, Num2Vec. We set up the changes in model accuracy when these three modules are not considered. Without focus loss, the relevancy evaluation branch cannot effectively overcome the problem that matching matrix  is very sparse, especially for SROIE dataset because of the little proportion of key-value pair. It can be seen that the relevancy evaluation branch (K-V matching) can improve the accuracy by a large margin. Without it, the performance of MatchVIE will decrease  4.09 \% in EPHOIE dataset beacause there is 85.62\% K-V ratio in it. In order to further verify the effectiveness of the relevancy evaluation branch, we give some prediction results of the MatchVIE model on whether removing the relevancy evaluation branch or not.  Note that when the relevancy evaluation branch is removed, our method becomes sequence labeling methods by relying on the  prediction of entity recognition branch.  From Figure 3,  it can be seen that sequence labeling methods are not effective enough for distinguishing numeric semantic categories or some ambiguous texts. On the contrary, our MatchVIE can effectively distinguish these categories by introducing the correlation between named entities. Besides, after employing Num2Vec, the model can achieve a more stable results with extra accuracy improvements.  In addition, we collect training loss and plot the loss curves. From Figure 4,  it can be seen that Num2Vec can help the model converge more smoothly. 

\subsection{Explorations of Network Architecture}
For the relevancy evaluation branch, we evaluate network architecture with different structures.

\paragraph{Comparisons of Different Mapping Methods.}  We try two methods to map a key to a certain category.  One is the lookup table given all possible keys for each category (e.g, the lookup table for the keys of the category `Total' can be `Total Sales', `Total', `Total Amount', `Total (RM)'),  and the other is based on semantic similarity, detailed in  section 3.4.   As shown in Table 3,  the lookup method is slightly better than semantics based method in EPHOIE dataset. However, the semantics based methods can flexibly customize categories and they don't need to build the lookup table in advance. Either method can be selected based on the actual situation.

\begin{table*}[]
		\setlength{\tabcolsep}{0.6mm}{
	\begin{tabular}{lccccccccccc}
		\hline
		\multirow{3}{*}{\textbf{Method}} & \multicolumn{11}{c}{\textbf{Entities}}                                                                                                                                                                                                                                                                                                                                                                               \\ \cline{2-12} 
		& \multirow{2}{*}{Subject} & \multirow{2}{*}{Test Time} & \multirow{2}{*}{Name} & \multirow{2}{*}{School} & \multirow{2}{*}{EXAM NO} & \multirow{2}{*}{Seat  NO} & \multirow{2}{*}{Class} & \multirow{2}{*}{STU NO} & \multirow{2}{*}{Grade} & \multirow{2}{*}{Score} & \multirow{2}{*}{\textbf{F1(\%)}} \\
		&                                   &                                     &                                &                                  &                                               &                                        &                                 &                                          &                                 &                                 &                              \\ \hline
		LSTM-CRF\cite{lample2016neural}              & 98.51                             & \textbf{100.0}                               & 98.87                          & 98.80                            & 75.86                                         & 72.73                                  & 94.04                           & 84.44                                    & 98.18                           & 69.57                           & 89.10                         \\

			(\cite{liu2019graph})               & 98.18                             & \textbf{100.0}                               & 99.52                          & \textbf{100.0}                            & 88.17                                         & 86.00                                  & 97.39                           & 80.00                                    & 94.44                           & 81.82                           & 92.55                        \\
		GraphIE ( \cite{qian2019graphie})       & 94.00                             & \textbf{100.0}                               & 95.84                          & 97.06                            & 82.19                                         & 84.44                                  & 93.07                           & 85.33                                    & 94.44                           & 76.19                           & 90.26                        \\
		TRIE  \cite{zhang2020trie}          & 98.79                             & \textbf{100.0}                               & 99.46                          & 99.64                            & 88.64                                         & 85.92                                  & 97.94                           & 84.32                                    & 97.02                           & 80.39                           & 93.21                        \\
			\textbf{MatchVIE(Ours)}                             & \textbf{99.78}                             & \textbf{100.0}                               & \textbf{99.88}                          & 98.57                            & \textbf{94.21}                                         & \textbf{93.48}                                  & \textbf{99.54}                           & \textbf{92.44}                                    & \textbf{98.35}                           & \textbf{92.45}                           & \textbf{96.87}   \\\hline                    
	\end{tabular}}
	\caption{  Experiment results on EPHOIE datasets. Standard F1-score (F1) are employed as evaluation metrics. LayoutLM only has pre-training models on English  data and  doesn't   work on EPHOIE which is in Chinese.  NO: Number, STU:Student, EXAM: Examination.}

\label{res1}
\end{table*}

\begin{figure}[t]

	\centering

	\includegraphics[width=1.0\columnwidth]{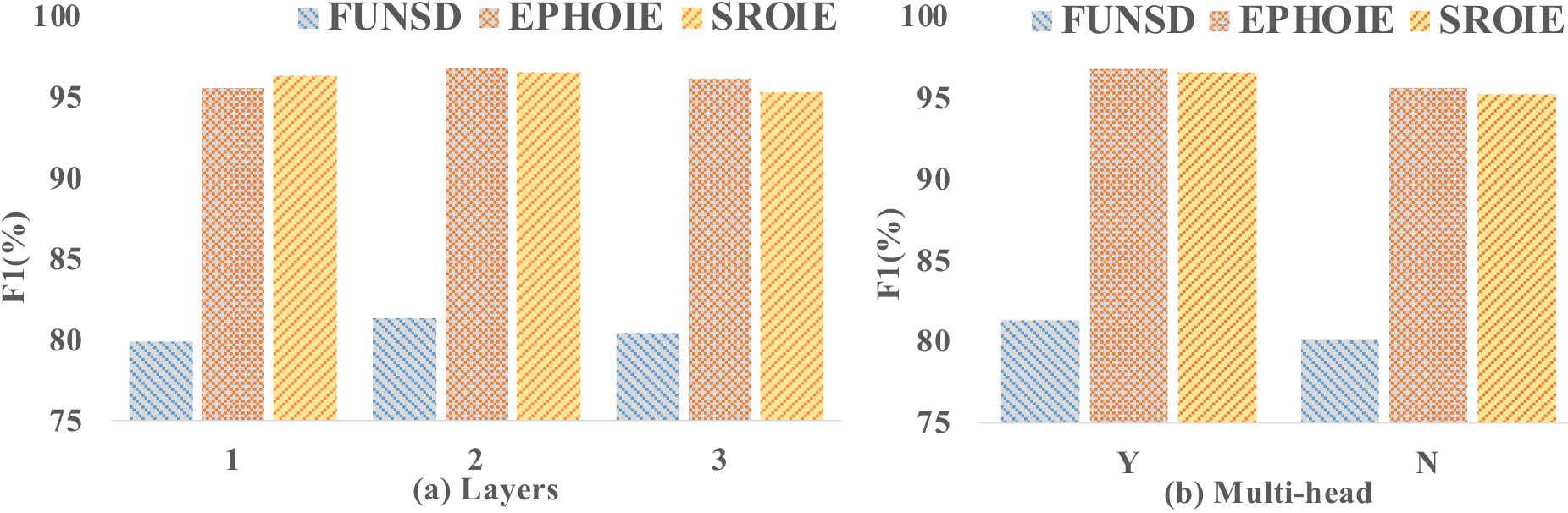}	
		\caption{Impacts of layers and heads in the relevancy evaluation branch.  The impacts of  number of layers (left). The  impacts of whether using muti-head or not (right).} 
	\label{lossnum2vec2}
\end{figure}

\paragraph{Network Architecture with Different Settings.} For the GNN in relevancy evaluation branch, we explore the appropriate network architecture with different settings.  By default, we use 8 heads for the multi-head attention. As for the layer number, the performance of two layers is better than that of one layer, as shown  in  Figure 5 (a). However, when three layers are used, the additional layer does not lead an evident gain. The results in Figure 5 (b) reveals that the multi-head attention can improve the accuracy to some extent.

\subsection{Comparisons with State-of-the-arts}
We compare our proposed model with the following several  methods: (1)  GraphIE  \cite{qian2019graphie}  is a VIE framework that  learns local and non-local contextual representations by graph structure. (2) Liu et al.  \cite{liu2019graph}  introduces a GNN model to combine global layout information into  BiLSTM-CRF. (3) LayoutLM \cite{xu2019layoutlm}  is an effective pre-training method with text and layout information.  (4) TRIE \cite{zhang2020trie} is a end-to-end method that combines visual and textual features for VIE. We
re-implement them based on the original papers or source
codes.

\paragraph{Results on EPHOIE Dataset.} The layout of images in EPHOIE dataset is complex and there are multiple types of entities, which are  easy to confuse. Most of these entities are numbers and do not contain semantics  (e.g, `EXAM NO', `Seat NO', `STU NO' and `Score' ). As shown in Table 4, to better illustrate the complementarity of our proposed MatchVIE with other methods, we give the F1-Score of all categories. Sequence labeling methods conduct sequence labeling on the documents where  text segments are concatenated from left to right and from top to bottom. This causes the determination of these numeric semantic categories to rely on the pre-organized order. Mainly owing to the strong correlation between entities, our MatchVIE can bypass the recognitions to various semantics, and simply focuses on the strong relevancy between entities. It achieves best  results and outperforms the state-of-the-art methods by significant margins (3.66\% than TRIE), especially in numeric semantic categories or some ambiguous texts. 

\begin{table}
	\centering
	\setlength{\tabcolsep}{2.0mm}{
		\begin{tabular}{lccccc}
			\hline
			\multirow{2}{*}{\textbf{Method}}    & \textbf{FUNSD}              & \textbf{} & \textbf{SROIE}        \\ \cline{2-2} \cline{4-4} 
			& \textbf{F1(\%)}  & \textbf{} & \textbf{F1(\%)} \\ \hline
			LSTM-CRF\cite{lample2016neural}                                       & 62.13                              &           & 90.85                 \\

			GraphIE \cite{qian2019graphie}                                & 72.12                                &           & 94.46                 \\
			(\cite{liu2019graph})                                                       & 72.43                                 &           & 95.10                 \\ 
			LayoutLM \cite{xu2019layoutlm}  †                           & 79.27                                   &           & 95.24                 \\
			
			TRIE \cite{zhang2020trie}                                                              & 78.86                              &           & 96.18                 \\
			
			\textbf{MatchVIE(Ours)}                                         & \textbf{81.33}              & \textbf{} & \textbf{96.57}        \\ \hline
	\end{tabular}}
	\caption{  Experiment results on FUNSD dataset and SROIE dataset. † indicates the results is reported in [Xu et al. 2019]}

	\label{res1}
\end{table}

\paragraph{Results on FUNSD Dataset.}  As shown in Table \ref{res1}, some multimodal methods, such as GraphIE and Liu's methods, achieve high performance by introducing layout information. Meantime, LayoutLM  achieves a better performance of 79.27\% when using the text, layout and image information at the same time. Our method  outperforms the state-of-the-art results by 2.06\% due to the introduction of entity relationship.

\paragraph{Results on SROIE Dataset.} The results of  experiments on SROIE datasets are shown in Table \ref{res1}. Our MatchVIE model  achieve a F1-score of 96.57\%, which is  better than the first place in SOTA methods. In this dataset, the location of these two categories of `Total' and `Date' are more flexible compared to other categories, and they are easily confused with other categories, as shown in Figure 3. Our method can effectively distinguish these two categories by combining neighbor relationships.  Although the relatively little proportion of key-value (54.33\%), our method has achieved competitive results.

\section{Conclusion}
 In this paper, we propose a novel MatchVIE model to extract the information from document images.  Through the key-value matching, the proposed MatchVIE can bypass the recognition to various semantics, and simply focuses on the strong relevancy between keys and values which has been guaranteed by the document layout or semantic information. Comprehensive experiments  show the benefit of learning the  key-value matching relationship explicitly.  We plan on extending this work to transfer the layout information between document images to achieve One-shot learning.
 \section*{Acknowledgments}
This research is supported in part by NSFC (Grant No.: 61936003, 61771199),  GD-NSF (no.2017A030312006).
 
\bibliographystyle{named}
\bibliography{ijcai21}

\begin{thebibliography}{}

\bibitem[\protect\citeauthoryear{Boro{\c{s}} \bgroup \em et al.\egroup
  }{2020}]{borocs2020comparison}
Emanuela Boro{\c{s}}, Ver{\'o}nica Romero, Martin Maarand, and Zenklov{\'a}.
\newblock A comparison of sequential and combined approaches for named entity
  recognition in a corpus of handwritten medieval charters.
\newblock In {\em ICFHR}, pages 79--84, 2020.

\bibitem[\protect\citeauthoryear{Boroumand \bgroup \em et al.\egroup
  }{2018}]{boroumand2018deep}
Mehdi Boroumand, Mo~Chen, and Jessica Fridrich.
\newblock Deep residual network for steganalysis of digital images.
\newblock {\em IEEE Trans. Inf.}, 14(5):1181--1193, 2018.

\bibitem[\protect\citeauthoryear{Carbonell \bgroup \em et al.\egroup
  }{2020}]{carbonell2020neural}
Manuel Carbonell, Alicia Forn{\'e}s, Mauricio Villegas, and Josep Llad{\'o}s.
\newblock A neural model for text localization, transcription and named entity
  recognition in full pages.
\newblock {\em Pattern Recognit Lett}, 2020.

\bibitem[\protect\citeauthoryear{Cheng \bgroup \em et al.\egroup
  }{2020}]{cheng2020one}
Mengli Cheng, Minghui Qiu, Xing Shi, Jun Huang, and Wei Lin.
\newblock One-shot text field labeling using attention and belief propagation
  for structure information extraction.
\newblock In {\em ACMmm}, pages 340--348, 2020.

\bibitem[\protect\citeauthoryear{Denk and Reisswig}{2019}]{denk2019bertgrid}
Timo~I Denk and Christian Reisswig.
\newblock {BERTgrid}: Contextualized embedding for 2d document representation
  and understanding.
\newblock In {\em NIPS}, 2019.

\bibitem[\protect\citeauthoryear{Devlin \bgroup \em et al.\egroup
  }{2019}]{devlin2019bert}
Jacob Devlin, Ming-Wei Chang, Kenton Lee, and Kristina Toutanova.
\newblock {BERT}: Pre-training of deep bidirectional transformers for language
  understanding.
\newblock In {\em NAACL}, pages 4171--4186, 2019.

\bibitem[\protect\citeauthoryear{Gal \bgroup \em et al.\egroup
  }{2020}]{gal2020cardinal}
Rinon Gal, Shai Ardazi, and Roy Shilkrot.
\newblock Cardinal graph convolution framework for document information
  extraction.
\newblock In {\em ACMmm}, pages 1--11, 2020.

\bibitem[\protect\citeauthoryear{Gui \bgroup \em et al.\egroup
  }{2019}]{gui2019lexicon}
Tao Gui, Yicheng Zou, Qi~Zhang, Minlong Peng, Jinlan Fu, Zhongyu Wei, and
  Xuan-Jing Huang.
\newblock A lexicon-based graph neural network for chinese {NER}.
\newblock In {\em EMNLP}, pages 1039--1049, 2019.

\bibitem[\protect\citeauthoryear{Guo \bgroup \em et al.\egroup
  }{2019}]{guo2019eaten}
He~Guo, Xiameng Qin, and Liu.
\newblock {EATEN}: Entity-aware attention for single shot visual text
  extraction.
\newblock In {\em ICDAR}, pages 254--259, 2019.

\bibitem[\protect\citeauthoryear{He \bgroup \em et al.\egroup
  }{2016}]{he2016deep}
Kaiming He, Xiangyu Zhang, Shaoqing Ren, and Jian Sun.
\newblock Deep residual learning for image recognition.
\newblock In {\em CVPR}, pages 770--778, 2016.

\bibitem[\protect\citeauthoryear{He \bgroup \em et al.\egroup
  }{2017}]{he2017mask}
Kaiming He, Georgia Gkioxari, Piotr Doll{\'a}r, and Ross Girshick.
\newblock {Mask RCNN}.
\newblock In {\em CVPR}, pages 2961--2969, 2017.

\bibitem[\protect\citeauthoryear{Huang \bgroup \em et al.\egroup
  }{2017}]{huang2017densely}
Gao Huang, Zhuang Liu, Laurens Van Der~Maaten, and Kilian~Q Weinberger.
\newblock Densely connected convolutional networks.
\newblock In {\em CVPR}, pages 4700--4708, 2017.

\bibitem[\protect\citeauthoryear{Huang \bgroup \em et al.\egroup
  }{2019}]{huang2019icdar2019}
Z~Huang, K~Chen, J~He, X~Bai, D~Karatzas, S~Lu, and CV~Jawahar.
\newblock {ICDAR}2019 competition on scanned receipt ocr and information
  extraction.
\newblock In {\em ICDAR}, pages 1516--1520, 2019.

\bibitem[\protect\citeauthoryear{Hwang \bgroup \em et al.\egroup
  }{2019}]{hwang2019post}
Wonseok Hwang, Seonghyeon Kim, Minjoon Seo, Jinyeong Yim, Seunghyun Park,
  Sungrae Park, Junyeop Lee, Bado Lee, and Hwalsuk Lee.
\newblock Post-ocr parsing: building simple and robust parser via bio tagging.
\newblock In {\em NIPS}, 2019.

\bibitem[\protect\citeauthoryear{Jaume \bgroup \em et al.\egroup
  }{2019}]{jaume2019funsd}
Guillaume Jaume, Hazim~Kemal Ekenel, and Jean-Philippe Thiran.
\newblock Funsd: A dataset for form understanding in noisy scanned documents.
\newblock In {\em ICDARW}, volume~2, pages 1--6, 2019.

\bibitem[\protect\citeauthoryear{Jiang \bgroup \em et al.\egroup
  }{2019a}]{jiang2019semi}
Bo~Jiang, Ziyan Zhang, Doudou Lin, Jin Tang, and Bin Luo.
\newblock Semi-supervised learning with graph learning-convolutional networks.
\newblock In {\em CVPR}, pages 11313--11320, 2019.

\bibitem[\protect\citeauthoryear{Jiang \bgroup \em et al.\egroup
  }{2019b}]{jiang2019integrating}
Zhaohui Jiang, Zheng Huang, Yunrui Lian, Jie Guo, and Weidong Qiu.
\newblock Integrating coordinates with context for information extraction in
  document images.
\newblock In {\em ICDAR}, pages 363--368, 2019.

\bibitem[\protect\citeauthoryear{Katti \bgroup \em et al.\egroup
  }{2018}]{katti2018chargrid}
Anoop~Raveendra Katti, Christian Reisswig, and Guder.
\newblock Chargrid: Towards understanding 2d documents.
\newblock In {\em EMNLP}, pages 4459--4469, 2018.

\bibitem[\protect\citeauthoryear{Kipf and Welling}{2016}]{kipf2016semi}
Thomas~N Kipf and Max Welling.
\newblock Semi-supervised classification with graph convolutional networks.
\newblock In {\em ICLR}, 2016.

\bibitem[\protect\citeauthoryear{Lample \bgroup \em et al.\egroup
  }{2016}]{lample2016neural}
Guillaume Lample, Miguel Ballesteros, and Subramanian.
\newblock Neural architectures for named entity recognition.
\newblock In {\em NAACL}, pages 260--270, 2016.

\bibitem[\protect\citeauthoryear{Lin \bgroup \em et al.\egroup
  }{2017}]{lin2017focal}
Tsung-Yi Lin, Priya Goyal, Ross Girshick, Kaiming He, and Piotr Doll{\'a}r.
\newblock Focal loss for dense object detection.
\newblock In {\em ICCV}, pages 2980--2988, 2017.

\bibitem[\protect\citeauthoryear{Liu \bgroup \em et al.\egroup
  }{2019}]{liu2019graph}
Xiaojing Liu, Feiyu Gao, Qiong Zhang, and Huasha Zhao.
\newblock Graph convolution for multimodal information extraction from visually
  rich documents.
\newblock In {\em NAACL}, pages 32--39, 2019.

\bibitem[\protect\citeauthoryear{Qian \bgroup \em et al.\egroup
  }{2019}]{qian2019graphie}
Yujie Qian, Enrico Santus, Zhijing Jin, Jiang Guo, and Regina Barzilay.
\newblock {GraphIE}: A graph-based framework for information extraction.
\newblock In {\em NAACL}, 2019.

\bibitem[\protect\citeauthoryear{SANG}{1999}]{sang1999representing}
EFTK SANG.
\newblock Representing text chunks.
\newblock In {\em EACL}, pages 173--179, 1999.

\bibitem[\protect\citeauthoryear{Veli{\v{c}}kovi{\'c} \bgroup \em et al.\egroup
  }{2017}]{velivckovic2017graph}
Petar Veli{\v{c}}kovi{\'c}, Guillem Cucurull, Arantxa Casanova, Adriana Romero,
  Pietro Lio, and Yoshua Bengio.
\newblock Graph attention networks.
\newblock In {\em ICLR}, 2017.

\bibitem[\protect\citeauthoryear{Wang \bgroup \em et al.\egroup
  }{2021}]{wangjia}
J~Wang, C~Liu, L~Jin, and G~Tang.
\newblock Towards robust visual information extraction in real world: New
  dataset and novel solution.
\newblock In {\em AAAI}, 2021.

\bibitem[\protect\citeauthoryear{Xu \bgroup \em et al.\egroup
  }{2020}]{xu2019layoutlm}
Yiheng Xu, Minghao Li, Lei Cui, Shaohan Huang, Furu Wei, and Ming Zhou.
\newblock {LayoutLM}: Pre-training of text and layout for document image
  understanding.
\newblock {\em KDD}, 2020.

\bibitem[\protect\citeauthoryear{Yu \bgroup \em et al.\egroup
  }{2020}]{yu2020pick}
Wenwen Yu, Ning Lu, Xianbiao Qi, Ping Gong, and Rong Xiao.
\newblock {PICK}: Processing key information extraction from documents using
  improved graph learning-convolutional networks.
\newblock {\em ICPR}, 2020.

\bibitem[\protect\citeauthoryear{Zhang \bgroup \em et al.\egroup
  }{2020}]{zhang2020trie}
Peng Zhang, Yunlu Xu, Zhanzhan Cheng, Shiliang Pu, Jing Lu, Liang Qiao, Yi~Niu,
  and Fei Wu.
\newblock {TRIE}: End-to-end text reading and information extraction for
  document understanding.
\newblock {\em ACMmm}, 2020.

\end{thebibliography}

\end{document}